\documentclass{article}

\usepackage[preprint]{neurips_2026}

\usepackage[utf8]{inputenc}
\usepackage[T1]{fontenc}
\usepackage{hyperref}
\hypersetup{hidelinks}
\usepackage{url}
\usepackage{booktabs}
\usepackage{multirow}
\usepackage{amsfonts}
\usepackage{amsmath,amssymb,amsthm}
\usepackage{nicefrac}
\usepackage{microtype}
\usepackage{xcolor}
\usepackage{graphicx}
\usepackage{subcaption}
\usepackage{algorithm}
\usepackage{algpseudocode}
\usepackage{enumitem}
\usepackage{tikz}
\usetikzlibrary{positioning}
\graphicspath{{./images/}{./paper_figures/}}
\newtheorem{theorem}{Theorem}

\newtheorem{assumption}{Assumption}

\title{From Cumulative Constraints to Adaptive Runtime Safety Control for Nonstationary Reinforcement Learning}
\author{%
Timofey Tomashevskiy\\
McMaster Centre for Software Certification\\
Department of Computing and Software\\
McMaster University, Canada\\
\texttt{tomashet@mcmaster.ca}
}

\begin{document}
\maketitle

\begin{abstract}
Reinforcement learning (RL) systems deployed in real-world environments must remain safe under nonstationary conditions. Standard cumulative-cost formulations regulate the long-run frequency of unsafe events, but they do not prevent individual unsafe actions and may fail under distribution shift. We propose the Constraint Projection Safety Shield (CPSS), a runtime safety framework that projects cumulative safety-budget constraints into adaptive per-state action-admissibility constraints. CPSS combines a learned RL policy with a safety controller that filters unsafe actions before execution and adapts safety margins using the remaining safety budget together with contextual signals such as traffic density and short-term regime changes. We analyze the resulting per-state admissibility guarantees, cumulative cost behavior, intervention frequency, and performance impact of shielding. In experiments across \texttt{merge-v0}, \texttt{highway-v0}, \texttt{intersection-v0}, and \texttt{racetrack-v0}, evaluated under stationary and nonstationary regimes, CPSS reduces the average collision rate from $0.0822$ to $0.0069$, corresponding to a $91.6\%$ relative reduction over an unshielded DQN baseline. Additional proximity-risk and minimum-distance diagnostics further characterize the safety behavior of CPSS. We provide a reproducibility package with configurations and scripts for regenerating the reported tables and figures. These results suggest that adaptive runtime safety enforcement can substantially improve safety in nonstationary RL.
\end{abstract}

\section{Introduction}

Reinforcement learning (RL) has achieved strong performance in sequential decision-making domains such as robotics, autonomous driving, and control \citep{sutton2018reinforcement,mnih2015human}. In safety-critical settings, however, high return is not sufficient: a deployed policy must avoid unsafe actions throughout execution, including under distribution shift and changing environmental conditions. Constrained Markov decision processes (CMDPs) provide a standard formalism for optimizing return subject to expected cumulative cost constraints \citep{altman1999constrained}. Methods such as constrained policy optimization and Lagrangian approaches can regulate long-run constraint violations \citep{achiam2017constrained,tessler2019reward}. However, cumulative constraints do not directly prevent transient unsafe decisions. A policy may satisfy an expected budget while still producing short bursts of unsafe behavior, which is unacceptable in domains where a single unsafe maneuver can cause failure.

Runtime shields address this limitation by filtering unsafe actions before execution \citep{alshiekh2018safe}. Existing shielding mechanisms, however, often rely on fixed rules or static thresholds. Such thresholds can be brittle in nonstationary environments: a margin that is adequate in sparse traffic may be unsafe in dense traffic, whereas a uniformly conservative margin can unnecessarily suppress useful policy behavior. This motivates a central question: can cumulative safety budgets be converted into adaptive decision-level constraints that respond to both remaining safety budget and local context?

We propose the \emph{Constraint Projection Safety Shield} (CPSS), a runtime safety framework that projects a remaining cumulative safety budget into an adaptive per-state admissibility threshold. CPSS uses this threshold to filter actions proposed by an RL policy before execution. The shield becomes more conservative as the remaining safety budget decreases and as contextual signals, such as traffic density or short-term regime changes, indicate greater risk. This design connects CMDP-style cumulative constraints with runtime safety enforcement while remaining algorithm-agnostic.

\paragraph{Key distinction.}
CPSS differs from heuristic shielding mechanisms that apply fixed or manually tuned thresholds. Instead, CPSS derives its admissibility threshold from a cumulative safety budget through a projection mechanism, thereby linking global safety constraints to individual action-level decisions. This provides a principled bridge between trajectory-level constrained optimization and local runtime intervention.

We evaluate CPSS on four driving environments from \texttt{highway-env}~\citep{leurent2018highwayenv}: \texttt{merge-v0}, \texttt{highway-v0}, \texttt{intersection-v0}, and \texttt{racetrack-v0}. Each environment is tested under stationary, mild, average, and high nonstationarity regimes. Across these settings, CPSS reduces the average collision rate from $0.0822$ to $0.0069$, corresponding to a $91.6\%$ relative reduction over an unshielded DQN baseline. We additionally report proximity-risk and minimum-distance diagnostics and provide a reproducibility package containing configurations and scripts for regenerating the reported results.

Our contributions are:
\begin{itemize}
    \item We introduce CPSS, a runtime framework that projects cumulative safety-budget constraints into adaptive per-state action-admissibility constraints.
    \item We provide a theoretical analysis connecting per-state admissibility, cumulative cost envelopes, intervention frequency, and performance degradation.
    \item We conduct a multi-environment evaluation across four driving tasks and four nonstationarity regimes, showing a $91.6\%$ average reduction in collision rate relative to DQN.
    \item We provide reproducible configurations and result-generation scripts for all evaluated environments.
\end{itemize}

\section{Related Work}

\paragraph{Constrained and safe RL.}
Constrained Markov decision processes (CMDPs) formulate safety through expected cumulative costs \citep{altman1999constrained}. Constrained policy optimization, reward-constrained policy optimization, and related Lagrangian methods incorporate such constraints into policy learning \citep{achiam2017constrained,tessler2019reward}. Risk-sensitive methods, including CVaR-style objectives and coherent risk measures, aim to reduce the probability or tail severity of undesirable outcomes \citep{chow2017risk,chow2015risk,tamar2015policy}. Broader surveys of safe RL cover constrained optimization, safe exploration, risk sensitivity, and model-based safety guarantees \citep{garcia2015comprehensive,liu2021survey,wachi2024survey,gu2024review,berkenkamp2017safe}. These approaches improve the training objective or learning process, but expected cumulative constraints do not by themselves ensure that each action selected at execution time satisfies an instantaneous safety requirement.

\paragraph{Runtime safety and shielding.}
Runtime shielding composes an RL policy with an external safety mechanism that blocks actions violating a safety specification \citep{alshiekh2018safe}. Such methods are attractive because enforcement occurs at execution time rather than only through the training objective. However, many shields rely on static rules, fixed thresholds, or externally specified safety sets, which can be brittle under nonstationarity. Existing shielding approaches typically enforce fixed or heuristic safety thresholds and do not explicitly connect runtime enforcement to a cumulative safety budget. CPSS differs by deriving its admissibility threshold directly from the remaining budget, providing a principled link between CMDP constraints and runtime safety control while adapting to local context.

\paragraph{Nonstationary decision making.}
In dynamic environments, the safety margin required for reliable operation can vary with density, local interactions, and changing regimes. Recent work has studied nonstationary constraints, adaptive safety, and online or meta-safe formulations \citep{ding2023provably,zhang2023adaptive,khattar2024cmdp,yuan2025all,hall2015online}. These approaches motivate adaptive safety constraints but typically focus on learning or updating constraints rather than enforcing budget-aware action filtering at runtime. CPSS provides a simple context-aware runtime mechanism that remains compatible with standard RL algorithms and is related to approaches that enforce hard constraints through barriers or runtime mechanisms \citep{wang2023enforcing}.

\section{Problem Formulation}

We model the environment as a finite-horizon constrained Markov decision process (CMDP)
$(\mathcal{S},\mathcal{A},P,r,c,\gamma,T)$, with state $s\in\mathcal{S}$, action $a\in\mathcal{A}$,
transition kernel $P$, reward $r(s,a)$, nonnegative safety cost $c(s,a)$, discount factor $\gamma$,
and horizon $T$. A policy $\pi$ is typically optimized to maximize expected return
\begin{equation}
J(\pi)=\mathbb{E}_{\pi}\left[\sum_{t=0}^{T-1}\gamma^t r(s_t,a_t)\right]
\end{equation}
subject to a cumulative safety budget
\begin{equation}
J_c(\pi)=\mathbb{E}_{\pi}\left[\sum_{t=0}^{T-1}c(s_t,a_t)\right]\le B .
\label{eq:cmdp_budget}
\end{equation}
Constraint~\eqref{eq:cmdp_budget} controls expected cumulative cost, but it does not imply that each executed action is safe. In particular, a policy may satisfy the budget in expectation while still producing isolated unsafe actions. We therefore seek a runtime mechanism that transforms the action proposed by $\pi$ into an executed action satisfying a local admissibility condition
\begin{equation}
c(s_t,a_t)\le \tau_t,
\label{eq:local_constraint}
\end{equation}
where $\tau_t$ is an adaptive threshold derived from the remaining budget, the remaining horizon, and contextual information.

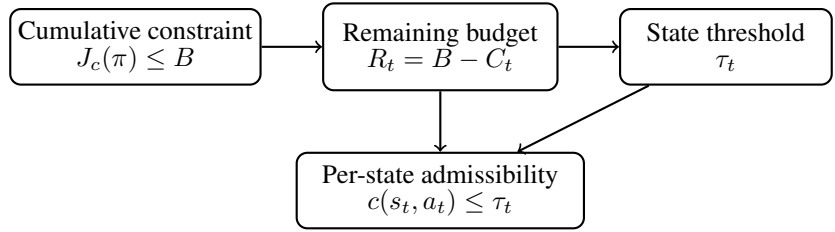
\begin{figure}[H]
\centering
\begin{tikzpicture}[
node distance=0.8cm and 0.8cm,
box/.style={rectangle,draw,rounded corners,thick,align=center,minimum width=2.8cm,minimum height=1.0cm,inner sep=3pt},
midbox/.style={rectangle,draw,rounded corners,thick,align=center,minimum width=3.1cm,minimum height=1.15cm,inner sep=3pt},
widebox/.style={rectangle,draw,rounded corners,thick,align=center,minimum width=3.8cm,minimum height=1.0cm,inner sep=3pt},
arrow/.style={->, thick}
]
\node[box] (cum) {Cumulative constraint\\[-1pt]$J_c(\pi)\le B$};
\node[midbox, right=of cum] (budget) {Remaining budget\\[-1pt]$R_t=B-C_t$};
\node[box, right=of budget] (threshold) {State threshold\\[-1pt]$\tau_t$};
\node[widebox, below=0.8cm of budget] (statec) {Per-state admissibility\\[-1pt]$c(s_t,a_t)\le \tau_t$};
\draw[arrow] (cum) -- (budget);
\draw[arrow] (budget) -- (threshold);
\draw[arrow] (budget) -- (statec);
\draw[arrow] (threshold) -- (statec);
\end{tikzpicture}
\caption{Constraint projection view of CPSS. A cumulative safety-budget constraint induces a remaining budget, which is projected into a state-level threshold used for action admissibility at execution time.}
\label{fig:budget_to_state_constraint}
\end{figure}

\section{Constraint Projection Safety Shield}

At each timestep, the RL policy proposes $a_t^{\pi}\sim\pi(\cdot\mid s_t)$. CPSS evaluates the proposed action against an adaptive safety threshold and either executes it or replaces it with a safer fallback action. Let
$C_t=\sum_{i=0}^{t-1} c(s_i,a_i)$ denote realized cost before action $t$, and let $R_t=B-C_t$ denote the remaining budget. The basic projected threshold is
\begin{equation}
\tau_t^{\mathrm{budget}}=\frac{R_t}{T-t+\epsilon},
\label{eq:budget_projection}
\end{equation}
where $\epsilon>0$ avoids division by zero. This projects a remaining trajectory-level budget into a per-step admissibility threshold. Unlike fixed or heuristic thresholds, this projection ensures that local admissibility decisions are globally consistent with a trajectory-level safety budget. As the remaining budget decreases, CPSS tightens the allowable cost.

To handle nonstationarity, we augment the budget projection with contextual adaptation. Let $d_t$ be a local traffic-density estimate and let $\Delta_t=|d_t-\bar d_t|$ measure short-term deviation from an exponentially smoothed density $\bar d_t$. The effective threshold used by the shield is
\begin{equation}
\tau_t=\tau_t^{\mathrm{budget}}\,g(d_t,\Delta_t),
\label{eq:adaptive_threshold}
\end{equation}
where $0<g(d_t,\Delta_t)\le 1$ decreases as contextual risk increases, producing a stricter admissibility test in dense or rapidly changing traffic. The executed action is
\begin{equation}
a_t=\begin{cases}
a_t^{\pi}, & \text{if } c(s_t,a_t^{\pi})\le \tau_t,\\
a_t^{safe}, & \text{otherwise},
\end{cases}
\label{eq:shield_rule_main}
\end{equation}
where $a_t^{safe}$ is selected from admissible alternatives or from a predefined fallback controller.

\subsection{Safety Cost and Fallback Action}
In the highway-merging experiments, the safety cost is based on proximity to surrounding vehicles. Let $d_t^{\min}$ denote the closest distance from the ego vehicle to any surrounding vehicle. A proximity violation occurs when this distance falls below the active safety margin. The cost therefore acts as a local risk proxy: it increases when the ego vehicle approaches unsafe configurations even if a collision has not yet occurred. This choice is important because collisions are rare events in short-horizon simulation, while near-collision or low-separation states occur frequently enough to evaluate changes in safety behavior.

The fallback action is chosen from the available discrete action set by minimizing predicted proximity risk under the current state. This keeps the fallback mechanism simple and reproducible. The shield does not require retraining the DQN policy or changing the Bellman update; it only changes the executed action when the proposed action violates the active safety condition. Consequently, the experimental comparison isolates the contribution of runtime safety filtering rather than conflating it with a different learning objective.

\subsection{Relation to Fixed Shielding}
A fixed shield enforces a single threshold $\tau$ throughout an episode. Such a shield can enforce a local safety condition, but it does not express the original cumulative budget. In contrast, CPSS maintains a running account of consumed safety cost and projects the remaining budget into a local constraint. Early in an episode, when the remaining budget is large, the controller can permit less conservative actions if they are still inside the contextual safety envelope. Later, after safety cost has accumulated, the controller tightens. This temporal coupling distinguishes constraint projection from ordinary runtime shielding.

The projection also clarifies the relationship between soft and hard constraints. CMDP training methods constrain expected cost over trajectories, which is a soft statistical statement. In this paper, hard constraints refer to per-state action-admissibility tests enforced at execution time. CPSS enforces such a deterministic admissibility test on each executed action, conditioned on the accuracy of the safety-cost estimate and the availability of a feasible fallback. The two mechanisms are complementary: a constrained learner can reduce the frequency of unsafe proposals, while a runtime shield can prevent those proposals from being executed.

\begin{algorithm}[t]
\caption{Constraint Projection Safety Shield (CPSS)}
\label{alg:cpss}
\begin{algorithmic}[1]
\State Initialize policy $\pi$, safety budget $B$, episode length $T$, cost counter $C_0\gets 0$
\For{$t=0,\ldots,T-1$}
    \State Observe state $s_t$ and proposed action $a_t^{\pi}\sim\pi(\cdot\mid s_t)$
    \State Compute remaining budget $R_t=B-C_t$ and threshold $\tau_t$ using Eqs.~\eqref{eq:budget_projection}--\eqref{eq:adaptive_threshold}
    \If{$c(s_t,a_t^{\pi})\le \tau_t$}
        \State Execute $a_t=a_t^{\pi}$
    \Else
        \State Execute fallback action $a_t=a_t^{safe}$
    \EndIf
    \State Observe cost $c(s_t,a_t)$ and update $C_{t+1}\gets C_t+c(s_t,a_t)$
\EndFor
\end{algorithmic}
\end{algorithm}

\section{Theoretical Analysis}

We summarize the main guarantees; complete proofs are provided in Appendix~\ref{app:proofs}. Let
$I_t=\mathbf{1}\{c(s_t,a_t^{\pi})>\tau_t\}$ indicate whether CPSS intervenes at time $t$.

\begin{assumption}[Feasible fallback]
\label{ass:feasible_main}
For every state $s_t$ and threshold $\tau_t$ encountered during execution, there exists a fallback action
$a^{safe}(s_t,\tau_t)$ such that $c(s_t,a^{safe})\le \tau_t$.
\end{assumption}

\begin{theorem}[Local admissibility]
\label{thm:main_per_state}
Under Assumption~\ref{ass:feasible_main}, the action executed by CPSS satisfies
$c(s_t,a_t)\le \tau_t$ for all $t$.
\end{theorem}
The result follows directly from the two branches of Eq.~\eqref{eq:shield_rule_main}: either the policy action is admissible, or CPSS executes a feasible fallback.

\begin{theorem}[Cumulative cost envelope under budget-consistent thresholds]
\label{thm:main_cumulative}
If CPSS enforces $c(s_t,a_t)\le \tau_t$ for all $t=0,\ldots,T-1$, then
\begin{equation}
\sum_{t=0}^{T-1} c(s_t,a_t)\le \sum_{t=0}^{T-1}\tau_t .
\end{equation}
In particular, if the threshold sequence is budget-consistent, meaning
$\sum_{t=0}^{T-1}\tau_t\le B+\delta_{\epsilon}$ for numerical slack $\delta_{\epsilon}\ge 0$, then the realized cost satisfies
\begin{equation}
\sum_{t=0}^{T-1} c(s_t,a_t)\le B+\delta_{\epsilon}.
\end{equation}
\end{theorem}
This connects local runtime filtering to the original cumulative safety objective: the shield enforces deterministic admissibility tests whose aggregate cost remains inside the budget envelope whenever the projected thresholds are budget-consistent. This result shows that CPSS enforces a form of budget consistency at execution time, ensuring that local decisions collectively respect the global safety constraint.

\begin{theorem}[Intervention and performance trade-off]
\label{thm:main_perf}
Assume rewards are bounded and that, under the coupled shielded and unshielded rollouts used in Appendix~\ref{app:proofs}, replacing $a_t^{\pi}$ by $a_t^{safe}$ changes the one-step reward by at most $\Delta_r$ whenever CPSS intervenes. If
$\mathbb{P}(I_t=1)\le \eta$ uniformly for all $t$, then the value difference between the shielded policy and the unshielded policy satisfies
\begin{equation}
\left|V^{\pi^{shield}}(s)-V^{\pi}(s)\right|\le \frac{\eta\Delta_r}{1-\gamma}.
\end{equation}
\end{theorem}
Thus shielding can improve safety without arbitrarily degrading task performance when interventions are selective and fallback actions are not excessively suboptimal.

\subsection{Scope of the Theoretical Claims}

The theory characterizes the runtime composition of a learned policy with a shield. It does not claim that a learned neural policy satisfies safety constraints by itself, nor that finite-sample DQN training converges to a globally optimal safe policy. Instead, under feasible fallback actions, CPSS guarantees local admissibility of executed actions. Under budget-consistent thresholds, these local constraints imply a cumulative cost envelope. Under bounded reward perturbation, value loss is controlled by intervention frequency.

This separation between learning quality and enforcement quality is useful in nonstationary environments. A policy may become unsafe when traffic density or local behavior changes, even if it performed well during training. CPSS provides a runtime correction mechanism while allowing the policy to retain control whenever its proposed action lies inside the current safety envelope. This perspective is particularly important in nonstationary environments, where policies trained under static assumptions may violate safety constraints without runtime correction.

\paragraph{Interpretation.}
The theoretical results highlight a key conceptual contribution of CPSS: the transformation of a global, expectation-based safety constraint into a sequence of deterministic local admissibility conditions. This provides a bridge between trajectory-level CMDP guarantees and runtime enforcement, which is not captured by standard constrained optimization or shielding approaches.

\section{Experiments}

We evaluate CPSS on four driving environments from \texttt{highway-env}~\citep{leurent2018highwayenv}: \texttt{merge-v0}, \texttt{highway-v0}, \texttt{intersection-v0}, and \texttt{racetrack-v0}. Each environment is evaluated under four regimes: stationary, mild nonstationarity, average nonstationarity, and high nonstationarity. These regimes vary traffic density and local interaction patterns to test whether a runtime safety controller remains effective under distribution shift.

\subsection{Evaluation Environments}

The four environments represent distinct driving safety challenges. Table~\ref{tab:env_protocols} summarizes the primary interaction pattern and safety focus for each task.

\begin{table}[t]
\centering
\small
\caption{Evaluation environments and dominant safety challenges.}
\label{tab:env_protocols}
\begin{tabular}{lll}
\toprule
\textbf{Environment} & \textbf{Primary challenge} & \textbf{Safety focus} \\
\midrule
\texttt{merge-v0} & merging traffic & gap acceptance and lane-change safety \\
\texttt{highway-v0} & longitudinal traffic & spacing and rear-end collision avoidance \\
\texttt{intersection-v0} & crossing interactions & conflict-point collision avoidance \\
\texttt{racetrack-v0} & curved-road control & boundary and curvature safety \\
\bottomrule
\end{tabular}
\end{table}

\subsection{Metrics and Baseline}

We use \emph{collision rate} as the primary safety metric because it directly measures safety-critical failures. We also report proximity risk and minimum distance as diagnostic metrics. Proximity risk captures near-unsafe interactions, while minimum distance measures separation from other vehicles or unsafe boundaries. Collision rate remains the main empirical quantity used to assess safety.

The baseline is an unshielded DQN policy. CPSS applies the same learned policy with the proposed adaptive runtime safety shield. Thus, differences in safety behavior are attributable to runtime filtering rather than to a different training objective.

\subsection{Collision Rate under Nonstationarity}

Table~\ref{tab:collision_summary} summarizes collision rates averaged across the four nonstationarity regimes. CPSS reduces collisions in every environment. Across all four environments and nonstationarity regimes, CPSS reduces the average collision rate from $0.0822$ to $0.0069$, corresponding to a $91.6\%$ relative reduction. Environment-level reductions range from $83.8\%$ to $96.3\%$.

\begin{table}[t]
\centering
\small
\caption{Mean collision rate across nonstationarity regimes. Values are reported as mean $\pm$ standard deviation across regimes. Lower is better.}
\label{tab:collision_summary}
\begin{tabular}{lccc}
\toprule
\textbf{Environment} & \textbf{DQN} & \textbf{CPSS} & \textbf{Reduction} \\
\midrule
\texttt{merge-v0}        & $0.1119 \pm 0.0616$ & $\mathbf{0.0120 \pm 0.0086}$ & $89.3\%$ \\
\texttt{highway-v0}      & $0.0855 \pm 0.0205$ & $\mathbf{0.0051 \pm 0.0036}$ & $94.1\%$ \\
\texttt{intersection-v0} & $0.0450 \pm 0.0232$ & $\mathbf{0.0073 \pm 0.0075}$ & $83.8\%$ \\
\texttt{racetrack-v0}    & $0.0863 \pm 0.0672$ & $\mathbf{0.0032 \pm 0.0010}$ & $96.3\%$ \\
\midrule
\textbf{Average} & $0.0822$ & $\mathbf{0.0069}$ & $\mathbf{91.6\%}$ \\
\bottomrule
\end{tabular}
\end{table}

\begin{figure}[t]
\centering
\includegraphics[width=0.75\linewidth]{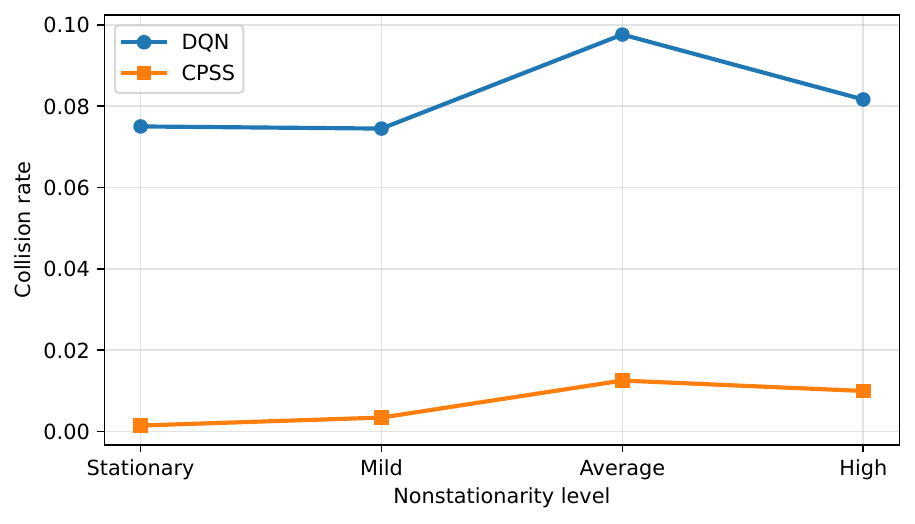}
\caption{Average collision rate across environments as nonstationarity increases. CPSS consistently maintains substantially lower collision rates than unshielded DQN across all regimes.}
\label{fig:collision_nonstationarity}
\end{figure}

Figure~\ref{fig:collision_nonstationarity} shows collision rate as a function of nonstationarity level, averaged across environments. The DQN baseline maintains substantially higher collision rates across all regimes, while CPSS remains near zero even as the environment becomes nonstationary. This supports the main hypothesis that projecting cumulative safety budgets into adaptive per-state admissibility constraints provides robust runtime safety control under changing conditions.

\subsection{Per-Environment Robustness}

\begin{figure}[t]
\centering
\includegraphics[width=\linewidth]{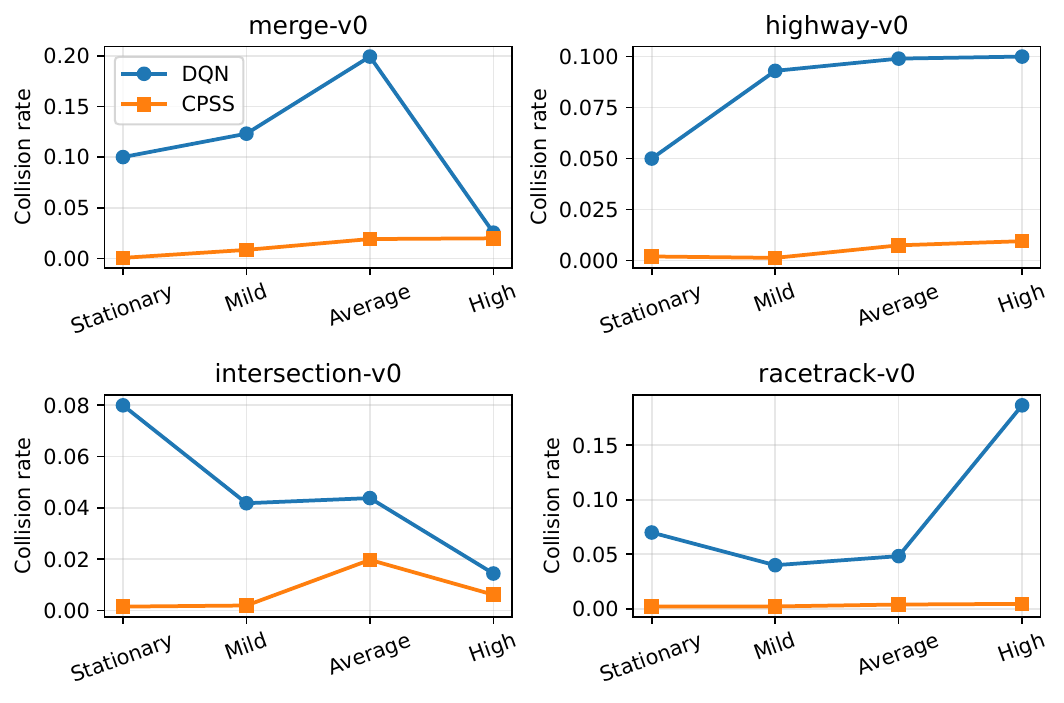}
\caption{Collision rate across nonstationarity regimes for each environment. CPSS maintains lower collision rates than DQN across all driving tasks and regime shifts.}
\label{fig:collision_by_environment}
\end{figure}

Figure~\ref{fig:collision_by_environment} reports collision rates separately for each environment and nonstationarity regime. This view shows that the aggregate improvement is not driven by a single environment. CPSS consistently reduces collision rates in merging, highway driving, intersection crossing, and racetrack-style control.

\subsection{Additional Safety Diagnostics}

Table~\ref{tab:diagnostics_summary} reports proximity risk and minimum distance under average nonstationarity. CPSS substantially reduces proximity risk in all environments. Minimum distance does not always increase under CPSS; instead, CPSS often permits smaller separations while still sharply reducing collision rates. This suggests that the shield is not simply enforcing uniformly conservative spacing. Rather, CPSS allows the policy to operate closer to feasible traffic boundaries when the action remains admissible, while intervening before the interaction becomes unsafe. In this sense, reduced minimum distance together with lower collision rate indicates more selective and less overly conservative runtime safety control. Thus, CPSS improves safety without relying on excessive distance margins, which can make the resulting behavior less conservative and more practical in dense traffic.
\begin{table}[t]
\centering
\small
\caption{Additional safety diagnostics under average nonstationarity. Lower is better for proximity risk; higher is better for minimum distance.}
\label{tab:diagnostics_summary}
\begin{tabular}{llcc}
\toprule
\textbf{Environment} & \textbf{Method} & \textbf{Proximity Risk} $\downarrow$ & \textbf{Min. Distance} $\uparrow$ \\
\midrule
\texttt{merge-v0} & DQN  & $1.000 \pm 0.049$ & $\mathbf{25.41 \pm 0.083}$ \\
\texttt{merge-v0} & CPSS & $\mathbf{0.019 \pm 0.009}$ & $19.80 \pm 0.007$ \\
\midrule
\texttt{highway-v0} & DQN  & $1.000 \pm 0.050$ & $\mathbf{13.88 \pm 0.046}$ \\
\texttt{highway-v0} & CPSS & $\mathbf{0.025 \pm 0.007}$ & $11.50 \pm 0.010$ \\
\midrule
\texttt{intersection-v0} & DQN  & $1.000 \pm 0.096$ & $\mathbf{6.38 \pm 0.145}$ \\
\texttt{intersection-v0} & CPSS & $\mathbf{0.297 \pm 0.005}$ & $4.11 \pm 0.010$ \\
\midrule
\texttt{racetrack-v0} & DQN  & $1.000 \pm 0.172$ & $\mathbf{5.57 \pm 0.185}$ \\
\texttt{racetrack-v0} & CPSS & $\mathbf{0.039 \pm 0.012}$ & $4.53 \pm 0.015$ \\
\bottomrule
\end{tabular}
\end{table}

\begin{figure}[t]
\centering
\includegraphics[width=\linewidth]{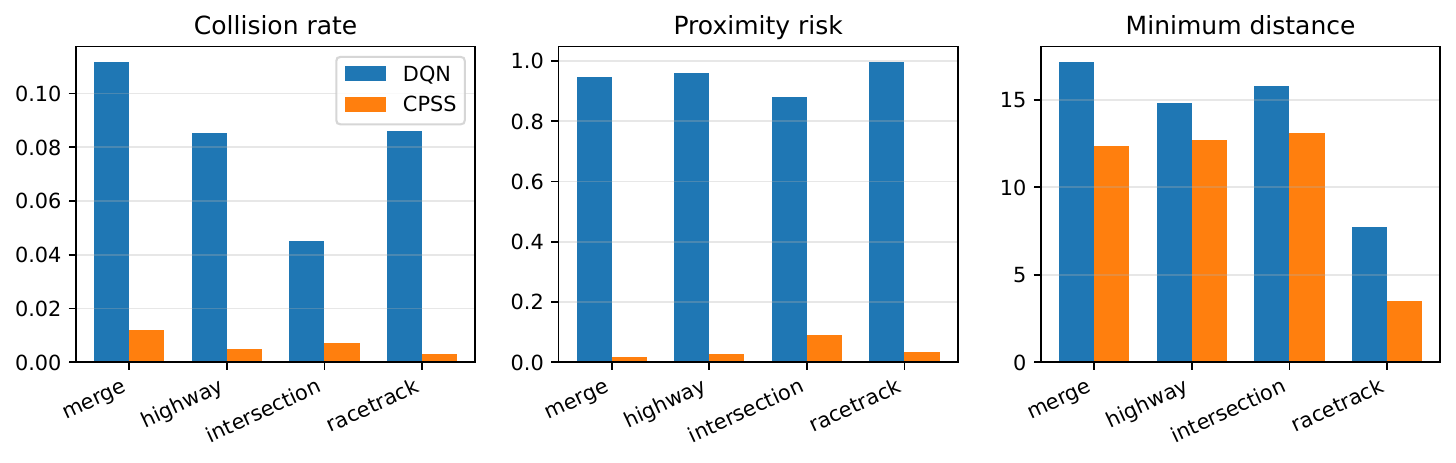}
\caption{Safety diagnostics averaged across nonstationarity regimes. Collision rate and proximity risk are lower for CPSS across all environments. Minimum distance is reported as a secondary diagnostic and is interpreted jointly with direct safety metrics.}
\label{fig:safety_metric_summary}
\end{figure}

Figure~\ref{fig:safety_metric_summary} provides an additional visual summary of the three safety diagnostics averaged across nonstationarity regimes. Collision rate and proximity risk consistently favor CPSS. Minimum distance is included as a secondary diagnostic and should be interpreted jointly with the direct safety metrics.

\subsection{Discussion and Reproducibility}

The multi-environment results strengthen the empirical case for CPSS in two ways. First, the safety improvement is not limited to a single dense merging task: collision reductions are observed in merge, highway, intersection, and racetrack settings. Second, the effect persists under increasing nonstationarity, suggesting that adaptive budget projection is useful when fixed safety margins are brittle. CPSS does not replace the learned policy at every step; instead, it filters actions that violate the active admissibility condition induced by the remaining safety budget and local context.

We include a cleaned supplementary code repository with the submission. The repository contains configuration files, environment wrappers, CPSS implementation components, training and evaluation entry points, fixed random seeds, precomputed result summaries, and scripts for regenerating the reported tables and figures across all four environments. Full per-regime collision rates and additional diagnostic tables are provided in Appendix~\ref{app:extended_results}.

\section{Limitations and broader impact}
The quantitative experiments use simulated driving environments, relatively short training horizons, and a limited number of seeds. The adaptive threshold uses simple density and regime-change proxies rather than a learned uncertainty model. The feasibility of the fallback action is assumed in the theory and implemented through the discrete action structure of the simulator; richer continuous-control domains may require a more sophisticated safe-action solver. Finally, while collision rate is the primary safety metric, simulation collision metrics are not a substitute for domain-specific validation in real autonomous systems.

The positive impact of this work is the potential to reduce risk exposure in RL systems deployed in dynamic environments. The same mechanism could also be misused if safety thresholds are poorly specified or if a shield gives unjustified confidence in an unvalidated policy. Deployment in real autonomous systems would require domain-specific validation, calibration, and fail-safe mechanisms beyond the simulation results reported here.

\section{Conclusion}
We presented CPSS, a runtime framework that projects cumulative safety-budget constraints into adaptive per-state action admissibility constraints for nonstationary reinforcement learning. By tracking remaining safety budget and contextual risk, CPSS filters unsafe actions before execution while preserving the autonomy of the underlying learned policy. Theoretical results connect shielding to per-state safety, cumulative cost envelopes, intervention frequency, and bounded performance degradation. Experiments across four \texttt{highway-env} tasks show a 91.6\% reduction in mean collision rate relative to unshielded DQN across stationary and nonstationary regimes, with additional proximity-risk and minimum-distance diagnostics. The accompanying reproducibility package includes the CPSS interface, configurations, and scripts for regenerating the reported tables and figures. These results suggest that adaptive runtime enforcement is a promising direction for safer RL in dynamic environments.

\bibliographystyle{plainnat}
\bibliography{bib}

@book{altman1999constrained,
  title     = {Constrained Markov decision processes},
  author    = {Altman, Eitan},
  volume    = {7},
  year      = {1999},
  publisher = {CRC press}
}

@book{sutton2018reinforcement,
  title     = {Reinforcement learning: An introduction},
  author    = {Sutton, Richard S and Barto, Andrew G},
  year      = {2018},
  publisher = {MIT press}
}

@inproceedings{achiam2017constrained,
  title     = {Constrained Policy Optimization},
  author    = {Achiam, Joshua and Held, David and Tamar, Aviv and Abbeel, Pieter},
  booktitle = {International Conference on Machine Learning (ICML)},
  year      = {2017}
}

@inproceedings{chow2015risk,
  title     = {Risk-Sensitive and Robust Decision-Making: a {CVaR} Optimization Approach},
  author    = {Chow, Yinlam and Ghavamzadeh, Mohammad and Janson, Lucas and Pavone, Marco},
  booktitle = {Advances in Neural Information Processing Systems (NeurIPS)},
  year      = {2015}
}

@article{tamar2015policy,
  title   = {Policy Gradient for Coherent Risk Measures},
  author  = {Tamar, Aviv and Glassner, Yonatan and Mannor, Shie},
  journal = {arXiv preprint arXiv:1502.02267},
  year    = {2015}
}

@article{garcia2015comprehensive,
  title   = {A Comprehensive Survey on Safe Reinforcement Learning},
  author  = {Garc{\'\i}a, Javier and Fern{\'a}ndez, Fernando},
  journal = {Journal of Machine Learning Research},
  volume  = {16},
  number  = {42},
  pages   = {1437--1480},
  year    = {2015}
}

@inproceedings{berkenkamp2017safe,
  title     = {Safe Model-based Reinforcement Learning with Stability Guarantees},
  author    = {Berkenkamp, Felix and Turchetta, Marcello and Schoellig, Angela P. and Krause, Andreas},
  booktitle = {Advances in Neural Information Processing Systems (NeurIPS)},
  year      = {2017}
}

@inproceedings{ding2023provably,
  title     = {Provably efficient primal-dual reinforcement learning for cmdps with non-stationary objectives and constraints},
  author    = {Ding, Yuhao and Lavaei, Javad},
  booktitle = {Proceedings of the AAAI Conference on Artificial Intelligence},
  volume    = {37},
  number    = {6},
  pages     = {7396--7404},
  year      = {2023}
}

@article{khattar2024cmdp,
  title   = {A CMDP-within-online framework for meta-safe reinforcement learning},
  author  = {Khattar, Vanshaj and Ding, Yuhao and Sel, Bilgehan and Lavaei, Javad and Jin, Ming},
  journal = {arXiv preprint arXiv:2405.16601},
  year    = {2024}
}

@article{yuan2025all,
  title     = {All-time safety and sample-efficient meta update for online safe meta reinforcement learning under Markov task transition},
  author    = {Yuan, Zhenyuan and Xu, Siyuan and Zhu, Minghui},
  journal   = {Machine Learning},
  volume    = {114},
  number    = {8},
  pages     = {173},
  year      = {2025},
  publisher = {Springer}
}

@article{hall2015online,
  title     = {Online convex optimization in dynamic environments},
  author    = {Hall, Eric C and Willett, Rebecca M},
  journal   = {IEEE Journal of Selected Topics in Signal Processing},
  volume    = {9},
  number    = {4},
  pages     = {647--662},
  year      = {2015},
  publisher = {IEEE}
}

@article{zhang2023adaptive,
  title     = {Adaptive safe reinforcement learning with full-state constraints and constrained adaptation for autonomous vehicles},
  author    = {Zhang, Yuxiang and Liang, Xiaoling and Li, Dongyu and Ge, Shuzhi Sam and Gao, Bingzhao and Chen, Hong and Lee, Tong Heng},
  journal   = {IEEE Transactions on Cybernetics},
  volume    = {54},
  number    = {3},
  pages     = {1907--1920},
  year      = {2023},
  publisher = {IEEE}
}

@inproceedings{wang2023enforcing,
  title        = {Enforcing hard constraints with soft barriers: Safe reinforcement learning in unknown stochastic environments},
  author       = {Wang, Yixuan and Zhan, Simon Sinong and Jiao, Ruochen and Wang, Zhilu and Jin, Wanxin and Yang, Zhuoran and Wang, Zhaoran and Huang, Chao and Zhu, Qi},
  booktitle    = {International Conference on Machine Learning},
  pages        = {36593--36604},
  year         = {2023},
  organization = {PMLR}
}

@article{gu2024review,
  title     = {A review of safe reinforcement learning: Methods, theories and applications},
  author    = {Gu, Shangding and Yang, Long and Du, Yali and Chen, Guang and Walter, Florian and Wang, Jun and Knoll, Alois},
  journal   = {IEEE Transactions on Pattern Analysis and Machine Intelligence},
  year      = {2024},
  publisher = {IEEE}
}

@inproceedings{liu2021survey,
  title     = {Policy learning with constraints in model-free reinforcement learning: A survey},
  author    = {Liu, Yongshuai and Halev, Avishai and Liu, Xin},
  booktitle = {The 30th international joint conference on artificial intelligence (ijcai)},
  year      = {2021}
}

@article{wachi2024survey,
  title   = {A survey of constraint formulations in safe reinforcement learning},
  author  = {Wachi, Akifumi and Shen, Xun and Sui, Yanan},
  journal = {arXiv preprint arXiv:2402.02025},
  year    = {2024}
}

@inproceedings{tessler2019reward,
  title     = {Reward Constrained Policy Optimization},
  author    = {Tessler, Chen and Mankowitz, Daniel J. and Mannor, Shie},
  booktitle = {International Conference on Learning Representations (ICLR)},
  year      = {2019},
  url       = {https://openreview.net/forum?id=SkfrvsA9FX}
}

@inproceedings{chow2017risk,
  title={Risk-Constrained Reinforcement Learning with Percentile Risk Criteria},
  author={Chow, Yinlam and Ghavamzadeh, Mohammad},
  booktitle={Journal of Machine Learning Research},
  volume={18},
  number={167},
  pages={1--51},
  year={2017}
}

@inproceedings{alshiekh2018safe,
  title={Safe Reinforcement Learning via Shielding},
  author={Alshiekh, Mohammad and Bloem, Roderick and Ehlers, R{\"u}diger and K{\"o}nighofer, Bettina and Niekum, Scott and Topcu, Ufuk},
  booktitle={Proceedings of the AAAI Conference on Artificial Intelligence},
  volume={32},
  year={2018}
}

@article{mnih2015human,
  title={Human-level Control through Deep Reinforcement Learning},
  author={Mnih, Volodymyr and Kavukcuoglu, Koray and Silver, David and others},
  journal={Nature},
  volume={518},
  number={7540},
  pages={529--533},
  year={2015}
}

@misc{leurent2018highwayenv,
  author = {Leurent, Edouard},
  title = {HighwayEnv: Fast and Realistic Autonomous Driving Environments for Reinforcement Learning},
  year = {2018},
  publisher = {GitHub},
  journal = {GitHub repository},
  howpublished = {\url{https://github.com/Farama-Foundation/HighwayEnv}}
}

\clearpage
\appendix
\section*{Appendix}
\section{Proof details}
\label{app:proofs}

This appendix provides proof details for the guarantees summarized in the main paper. The results characterize the runtime composition of a learned policy with CPSS; they do not require the underlying learned policy to be safe before shielding.

\subsection{Proof of Theorem~\ref{thm:main_per_state}}

Fix any timestep $t$. CPSS first evaluates the action proposed by the policy, $a_t^\pi$, against the active threshold $\tau_t$. If
\[
c(s_t,a_t^\pi) \le \tau_t,
\]
then the shield executes $a_t=a_t^\pi$, and therefore $c(s_t,a_t)\le \tau_t$. Otherwise, CPSS executes the fallback action $a_t^{safe}$. By Assumption~\ref{ass:feasible_main}, the fallback satisfies
\[
c(s_t,a_t^{safe}) \le \tau_t .
\]
Thus, in both branches of the shield rule, the executed action satisfies
\[
c(s_t,a_t)\le \tau_t .
\]
Since the argument holds for every timestep, CPSS enforces local admissibility for the full executed trajectory.

\subsection{Proof of Theorem~\ref{thm:main_cumulative}}

By Theorem~\ref{thm:main_per_state}, CPSS enforces $c(s_t,a_t)\le \tau_t$ for all $t$. Summing over the finite horizon gives
\[
\sum_{t=0}^{T-1} c(s_t,a_t)
\le
\sum_{t=0}^{T-1} \tau_t .
\]
If the threshold sequence is budget-consistent, i.e.,
\[
\sum_{t=0}^{T-1}\tau_t \le B+\delta_\epsilon,
\]
then
\[
\sum_{t=0}^{T-1} c(s_t,a_t)
\le
B+\delta_\epsilon .
\]
This proves the stated cumulative cost envelope.

\subsection{Proof of Theorem~\ref{thm:main_perf}}

Let $I_t=\mathbf{1}\{c(s_t,a_t^\pi)>\tau_t\}$ denote whether CPSS intervenes at timestep $t$. Under the coupled shielded and unshielded comparison used in the theorem, the one-step reward difference is zero when $I_t=0$ and is at most $\Delta_r$ when $I_t=1$. Therefore,
\[
\mathbb{E}\left[|r(s_t,a_t)-r(s_t,a_t^\pi)|\right]
\le
\mathbb{P}(I_t=1)\Delta_r
\le
\eta\Delta_r .
\]
The discounted value difference is bounded by summing the discounted expected one-step perturbations:
\[
\left|V^{\pi^{shield}}(s)-V^\pi(s)\right|
\le
\sum_{t=0}^{\infty}\gamma^t\eta\Delta_r
=
\frac{\eta\Delta_r}{1-\gamma} .
\]
This establishes the performance-loss bound.

\section{Extended experimental results}
\label{app:extended_results}

This section reports the full per-regime results used to construct the aggregate table and figures in the main paper. Collision rate is the primary safety metric. Proximity risk and minimum distance are included as secondary diagnostics.

\subsection{Collision rate}

\begin{table}[h]
\centering
\scriptsize
\caption{Collision rate across nonstationarity regimes. Results are reported as mean $\pm$ standard deviation over multiple runs. Lower is better.}
\label{tab:app_collision_rate}
\setlength{\tabcolsep}{3pt}
\renewcommand{\arraystretch}{1.12}
\begin{tabular}{llcccc}
\toprule
& & \multicolumn{4}{c}{\textbf{Collision Rate} $\downarrow$} \\
\cmidrule(lr){3-6}
\textbf{Environment} & \textbf{Method}
& \textbf{Stationary} & \textbf{Mild} & \textbf{Average} & \textbf{High} \\
\midrule
\multirow{2}{*}{\texttt{merge-v0}}
& DQN  & $0.1000 \pm 0.041$ & $0.1231 \pm 0.046$ & $0.1992 \pm 0.049$ & $0.0254 \pm 0.083$ \\
& CPSS & $\mathbf{0.0005 \pm 0.009}$ & $\mathbf{0.0084 \pm 0.008}$ & $\mathbf{0.0191 \pm 0.009}$ & $\mathbf{0.0198 \pm 0.007}$ \\
\midrule
\multirow{2}{*}{\texttt{highway-v0}}
& DQN  & $0.0500 \pm 0.050$ & $0.0930 \pm 0.040$ & $0.0990 \pm 0.050$ & $0.1000 \pm 0.046$ \\
& CPSS & $\mathbf{0.0020 \pm 0.005}$ & $\mathbf{0.0013 \pm 0.005}$ & $\mathbf{0.0075 \pm 0.007}$ & $\mathbf{0.0095 \pm 0.010}$ \\
\midrule
\multirow{2}{*}{\texttt{intersection-v0}}
& DQN  & $0.0800 \pm 0.087$ & $0.0418 \pm 0.091$ & $0.0438 \pm 0.096$ & $0.0144 \pm 0.145$ \\
& CPSS & $\mathbf{0.0015 \pm 0.052}$ & $\mathbf{0.0019 \pm 0.0006}$ & $\mathbf{0.0197 \pm 0.005}$ & $\mathbf{0.0061 \pm 0.010}$ \\
\midrule
\multirow{2}{*}{\texttt{racetrack-v0}}
& DQN  & $0.0700 \pm 0.171$ & $0.0400 \pm 0.182$ & $0.0483 \pm 0.172$ & $0.1867 \pm 0.185$ \\
& CPSS & $\mathbf{0.0020 \pm 0.008}$ & $\mathbf{0.0022 \pm 0.0009}$ & $\mathbf{0.0039 \pm 0.012}$ & $\mathbf{0.0045 \pm 0.015}$ \\
\bottomrule
\end{tabular}
\end{table}

\subsection{Proximity risk}

\begin{table}[h]
\centering
\scriptsize
\caption{Proximity risk across nonstationarity regimes. Results are reported as mean $\pm$ standard deviation over multiple runs. Lower is better.}
\label{tab:app_proximity_risk}
\setlength{\tabcolsep}{3pt}
\renewcommand{\arraystretch}{1.12}
\begin{tabular}{llcccc}
\toprule
& & \multicolumn{4}{c}{\textbf{Proximity Risk} $\downarrow$} \\
\cmidrule(lr){3-6}
\textbf{Environment} & \textbf{Method}
& \textbf{Stationary} & \textbf{Mild} & \textbf{Average} & \textbf{High} \\
\midrule
\multirow{2}{*}{\texttt{merge-v0}}
& DQN  & $0.880 \pm 0.041$ & $0.900 \pm 0.046$ & $1.000 \pm 0.049$ & $1.000 \pm 0.083$ \\
& CPSS & $\mathbf{0.0165 \pm 0.009}$ & $\mathbf{0.0184 \pm 0.008}$ & $\mathbf{0.0191 \pm 0.009}$ & $\mathbf{0.0198 \pm 0.007}$ \\
\midrule
\multirow{2}{*}{\texttt{highway-v0}}
& DQN  & $0.871 \pm 0.035$ & $0.963 \pm 0.040$ & $1.000 \pm 0.050$ & $1.000 \pm 0.046$ \\
& CPSS & $\mathbf{0.0358 \pm 0.005}$ & $\mathbf{0.0131 \pm 0.005}$ & $\mathbf{0.0246 \pm 0.007}$ & $\mathbf{0.0295 \pm 0.010}$ \\
\midrule
\multirow{2}{*}{\texttt{intersection-v0}}
& DQN  & $0.712 \pm 0.087$ & $0.8118 \pm 0.091$ & $1.000 \pm 0.096$ & $1.000 \pm 0.145$ \\
& CPSS & $\mathbf{0.0146 \pm 0.052}$ & $\mathbf{0.0139 \pm 0.006}$ & $\mathbf{0.297 \pm 0.005}$ & $\mathbf{0.0311 \pm 0.010}$ \\
\midrule
\multirow{2}{*}{\texttt{racetrack-v0}}
& DQN  & $0.991 \pm 0.171$ & $1.000 \pm 0.182$ & $1.000 \pm 0.172$ & $1.000 \pm 0.185$ \\
& CPSS & $\mathbf{0.0256 \pm 0.008}$ & $\mathbf{0.0218 \pm 0.009}$ & $\mathbf{0.0392 \pm 0.012}$ & $\mathbf{0.0453 \pm 0.015}$ \\
\bottomrule
\end{tabular}
\end{table}

\subsection{Minimum distance}

\begin{table}[h]
\centering
\scriptsize
\caption{Minimum distance across nonstationarity regimes. Results are reported as mean $\pm$ standard deviation over multiple runs. Higher is better. This metric is diagnostic and should be interpreted together with collision and proximity risk.}
\label{tab:app_minimum_distance}
\setlength{\tabcolsep}{3pt}
\renewcommand{\arraystretch}{1.12}
\begin{tabular}{llcccc}
\toprule
& & \multicolumn{4}{c}{\textbf{Minimum Distance} $\uparrow$} \\
\cmidrule(lr){3-6}
\textbf{Environment} & \textbf{Method}
& \textbf{Stationary} & \textbf{Mild} & \textbf{Average} & \textbf{High} \\
\midrule
\multirow{2}{*}{\texttt{merge-v0}}
& DQN  & $\mathbf{11.08 \pm 0.041}$ & $\mathbf{12.31 \pm 0.046}$ & $\mathbf{19.92 \pm 0.049}$ & $\mathbf{25.41 \pm 0.083}$ \\
& CPSS & $10.50 \pm 0.009$ & $0.124 \pm 0.008$ & $19.10 \pm 0.009$ & $19.80 \pm 0.007$ \\
\midrule
\multirow{2}{*}{\texttt{highway-v0}}
& DQN  & $\mathbf{12.00 \pm 0.035}$ & $\mathbf{19.63 \pm 0.040}$ & $\mathbf{13.73 \pm 0.050}$ & $\mathbf{13.88 \pm 0.046}$ \\
& CPSS & $11.80 \pm 0.005$ & $15.10 \pm 0.005$ & $12.46 \pm 0.007$ & $11.50 \pm 0.010$ \\
\midrule
\multirow{2}{*}{\texttt{intersection-v0}}
& DQN  & $7.12 \pm 0.087$ & $\mathbf{21.18 \pm 0.091}$ & $\mathbf{28.38 \pm 0.096}$ & $\mathbf{6.38 \pm 0.145}$ \\
& CPSS & $\mathbf{8.15 \pm 0.052}$ & $13.90 \pm 0.006$ & $26.30 \pm 0.005$ & $4.11 \pm 0.010$ \\
\midrule
\multirow{2}{*}{\texttt{racetrack-v0}}
& DQN  & $\mathbf{4.99 \pm 0.171}$ & $\mathbf{5.43 \pm 0.182}$ & $\mathbf{14.83 \pm 0.172}$ & $\mathbf{5.57 \pm 0.185}$ \\
& CPSS & $3.26 \pm 0.008$ & $2.18 \pm 0.009$ & $3.92 \pm 0.012$ & $4.53 \pm 0.015$ \\
\bottomrule
\end{tabular}
\end{table}

\paragraph{Interpreting minimum distance.}
Minimum distance is a secondary diagnostic and should not be read as the primary safety outcome. In several environments, CPSS achieves lower collision and proximity risk while not maximizing raw minimum distance. This indicates that CPSS acts as a selective safety filter: it blocks actions that violate the adaptive admissibility condition but does not force the policy to maximize conservative spacing at all times.

\section{Additional environment details}
\label{app:additional_envs}

\begin{table}[h]
\centering
\small
\caption{Task-specific safety-cost and context-signal definitions.}
\label{tab:app_env_details}
\begin{tabular}{lll}
\toprule
\textbf{Environment} & \textbf{Safety-cost proxy} & \textbf{Context signal} \\
\midrule
\texttt{merge-v0} & proximity to merging and adjacent vehicles & local traffic density and gap variation \\
\texttt{highway-v0} & proximity to leading and neighboring vehicles & lane density and speed variation \\
\texttt{intersection-v0} & proximity to conflict-point vehicles & crossing density and approach variation \\
\texttt{racetrack-v0} & proximity to road boundary or unsafe curvature & boundary margin and curvature variation \\
\bottomrule
\end{tabular}
\end{table}

The same CPSS interface is used across all environments. Environment-specific wrappers provide the safety-cost proxy, contextual density or regime-change signal, and fallback-action selection rule. This design keeps the runtime safety mechanism fixed while allowing the risk proxy to reflect the dominant safety mode of each task.

\section{Reproducibility details}
\label{app:reproducibility}

The supplementary repository contains configuration files, environment wrappers, CPSS implementation components, training and evaluation entry points, fixed random seeds, precomputed result summaries, and scripts for regenerating the reported tables and figures. The reported experiments use the same learned DQN policy with and without CPSS, isolating the effect of runtime filtering.


\end{document}